# EpilNet: A Novel Approach to IoT based Epileptic Seizure Prediction and Diagnosis System using Artificial Intelligence

*\*Shivam Gupta[a], Virender Ranga[b], Priyansh Agrawal[b]*

[a]Department of Computer Science and Engineering, Indian Institute of Information Technology, Sonepat, Haryana, India (Mentor National Institute of Technology, Kurukshetra, Haryana, India)
[b]Department of Computer Engineering, National Institute of Technology, Kurukshetra, Haryana, India.

*\* Corresponding author: Shivam Gupta, shivi98g@gmail.com*

**ABSTRACT**

Epilepsy is one of the most occurring neurological diseases. The main characteristic of this disease is a frequent seizure, which is an electrical imbalance in the brain. It is generally accompanied by shaking of body parts and even leads (fainting). In the past few years, many treatments have come up. These mainly involve the use of anti-seizure drugs for controlling seizures. But in 70% of cases, these drugs are not effective, and surgery is the only solution when the condition worsens. So patients need to take care of themselves while having a seizure and be safe. Wearable electroencephalogram (EEG) devices have come up with the development in medical science and technology. These devices help in the analysis of brain electrical activities. EEG helps in locating the affected cortical region. The most important is that it can predict any seizure in advance on-site. This has resulted in a sudden increase in demand for effective and efficient seizure prediction and diagnosis systems. A novel approach to epileptic seizure prediction and diagnosis system "EpilNet" is proposed in the present paper. It is a one-dimensional (1D) convolution neural network. EpilNet gives the testing accuracy of 79.13% for five classes, leading to a significant increase of about 6-7% compared to related works. The developed Web API helps in bringing EpilNet into practical use. Thus, it is an integrated system for both patients and doctors. The system will help patients prevent injury or accidents and increase the efficiency of the treatment process by doctors in the hospitals.

## 1. Introduction

Epilepsy is a long lasting neurological disease that affects 70 million globally. It is not a newly discovered disease traced back to 4000 BC. The disease is not spread by direct contact with affected person. The main causes involve loss of oxygen to brain, or severe head injury, brain stroke and even can be due to genetic heredity. Epilepsy is mainly characterized by frequent seizures [12]. Seizure involves electrical imbalance in brain. These imbalances are accompanied with random shaking or involuntary movements of body parts [7].In past few decades with advancement in technology and medical science treatments are available for seizure control. Most of these involve use of anti-seizure drugs. But in 80% of cases these drugs are not effective in controlling frequency of seizures. Presently Electroencephalogram (EEG) is widely used for monitoring brain activities. EEG are mainly of two types. One of them is wearable EEG that can be placed on human scalp as shown in Fig.1. Other EEG is implanted in skull [8]. Wave signals from EEG electrodes helps in diagnosis of Epilepsy. They are also useful for predicting on-site time for a seizure. This all has created a sudden increase in demand for seizure prediction systems. These systems will help patient by generating alarms before seizure. This will prevent any kind of injury or accident due to uncertain seizure. Strong and long lasting seizures can even sometime lead to black out (fainting). Such systems will also help pregnant ladies to stabilize themselves before seizure. This will help in reducing miscarriages due to sudden falling down or hitting sharp objects nearby. Most of epilepsy cases are reported in childhood [3]. Small children with the development of such systems will get assistance. It will help them save themselves while physical activities like playing games, swimming etc. So in all there is a rising need and urge for some epilepsy seizure prediction and diagnosis system.

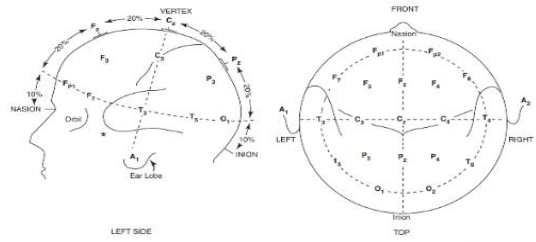

**Fig. 1.** Wearable EEG device on head scalp

Recently in February 2020 Liu et al. proposed 2D convolution neural network (CNN) that gave an accuracy of 64.5% for five classes [13]. In January 2020, Mao et al. came up with a combined CNN model with continuous wavelet transformation for five class classification of epileptic signal that gave an accuracy of 72.49% [14]. Abbasi et al. proposed a long short term memory (LSTM) based recurrent neural network (RNN) classifier that classified pre-ictal, ictal with an accuracy of 94% [1]. Cruz et al. in 2018 came up with a prototype embedded in gloves and proposed a standalone application that used motion to detect seizure and send short messaging service (SMS) alert [10]. Ranganathan et al. proposed a seizure diary in which patient has to manually record the seizure details [6]. Marzuki et al. proposed 'MyEpiPal' in which patients can manually add many details related to epilepsy [15]. Application is using inbuilt mobile accelerometer to detect motion during seizure.

Work has also been carried out in development of wearable EEG devices. Sogamoso et al. in 2018 came up with a graphene based electrode that can take EEG signal from brain [16]. The electrode is made up of carbon and is light in nature. It is wearable. Ahmed et al. came up with a wrist and chest wearable EEG device in 2017 [9]. In 2016, two different wearable EEG headset were proposed by Dzaferovic et al. and Myers et al. that could send signal to android application using Bluetooth [11, 4]. Pinho et al. proposed a wearable cap containing electrodes that can capture brain activities and send related EEG signals to IoT devices [5].

It is pertinent from the literature review that many recent works have come up with development of wearable and handy devices. These can be connected to automated systems using Internet of Things (IoT). Also it should be noted that most of the work revolve around involvement of patients by manually adding about the seizure details. In case of existing alarm predicting system there is a lot of scope for improvement in increasing the reliability, efficiency and usability. Also the most critical area left out in present works is development of an integrated automated system for patients and doctors that will help in prediction, diagnosis and analysis of seizure. Along with this a lot of scope of improvement is there in the development of better deep learning based classifiers. Most of the work revolves around converting the true nature of raw textual EEG data into two dimensional (2D) form. This destroys true the originality of data. Therefore if the true nature of data is preserved there are scope of improvement in efficiency of model.

Keeping these points in mind, in this paper a novel approach for integrated IoT based system is proposed. The system will be using a deep learning based neural model architecture 'EpilNet'. EpilNet is one dimensional (1D) model that preserves the raw textual nature of EEG data. The EpilNet will use the proposed algorithm for values of hyper parameters and dropout ratios. An API has also been proposed to bring EpilNet into practical use for integrated system. The Android based system will help in predicting onsite time quickly by use of API. The approach will not only help in keeping patients safe but also help doctors for timely treatment and efficient diagnosis at hospitals using proposed web based system.

## 2. Material and Method

The material and method used in present study are discussed below.

### 2.1. Dataset

Epileptic Seizure Recognition Dataset (ESRD) contains samples of patients affected by epilepsy. The dataset is prepared by Department of Epileptology, Bonn University. The dataset is free to use and is available on UCI Machine learning repository for research purposes [2]. The data is sampled and recorded at 173.61 Hz. The duration of each sample is around 23.6 seconds. These samples recorded are amplified using 128 channels. Samples are broadly divided into five main classes shown in Table 1. The dataset contains 11500 EEG samples. Each sample having a length of 178 and 1 label for class in numerical format with A-E as 1 to 5. The raw EEG signals plotted using matplotlib in python are shown in Fig. 2. It helps in better understanding of all classes.

**Table 1.** Description of dataset classes

| Label | Class Name | Remarks |
|---|---|---|
| A | Healthy Open Eyes | Person not suffering from Epilepsy |
| B | Healthy Closed Eyes | Person not suffering from Epilepsy |
| C | Inter-Ictal | Signals of unhealthy person that are neither pre-ictal nor ictal |
| D | Pre-Ictal | Signals of unhealthy person before seizure |
| E | Ictal | Signals of unhealthy person during a seizure |

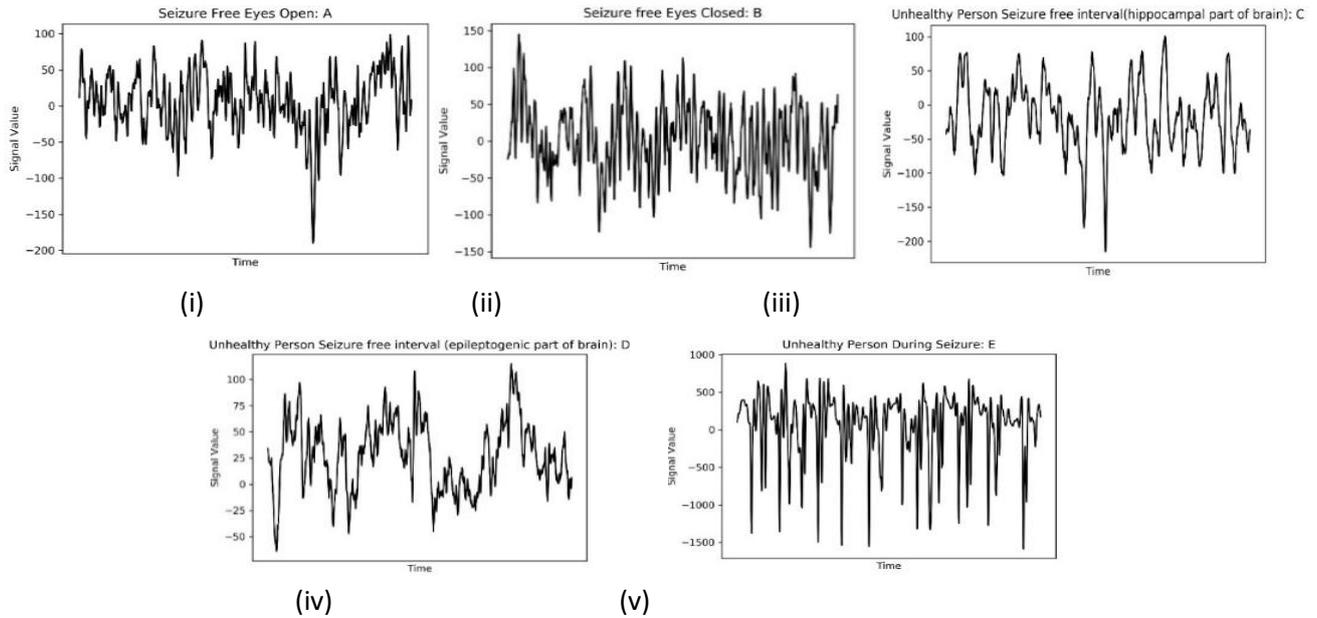

**Fig. 2.** Sample recording of EEG signal graph for (i) class A (ii) class B (iii) class C (iv) class D (v) class E

## 2.2. Data Preprocessing

In the present study for development of an integrated system, the dataset is divided and grouped together. The groups formed are significant for prediction or diagnosis of epilepsy. The grouped data will be used further for training of EpilNet. The dataset contains mainly raw EEG signal values for five classes. The classes are namely A to E. For development of prediction system to find on-site time of seizure. The data are grouped together as (AB), (D), (E).Thus the problem reduces to 3 class classification. For diagnosis and analysis by doctors all five classes are significant. These five classes are treated as separate groups. So problem becomes a five class classification for EpilNet.

## 2.3. EpilNet Model

EpilNet is mainly formed using basic block. The basic block comprises up of two 1D layers. These two layer are convolution 1D layer having filter of size 3 along with padding of one. The output feature map from this layer undergoes batch normalization. Basic block has additional skip connection to avoid vanishing gradient. Skip connection adds input feature map of basic block to output map.

Now to form complete architecture of EpilNet, the basic blocks are used. The raw textual EEG signal in form of 1D vector are given as input to EpilNet. This input undergoes a convolution operation of one dimension. The operation has a filter size of 7 with a stride of 2. The padding value of 3 is applied to maintain output feature map dimensions. The output feature map undergoes an activation and normalization. The activation function used is ReLU. So obtained, feature map now passes through a chain of basic blocks. Each chain of basic block is repeated (3, 3, 4, 4) times. The output map obtained after passing from basic block undergoes an average pooling.

The EpilNet is then converted into fully connected network. At last of fully connected network the softmax function is applied to predict label for input signals. EpilNet makes a total of 32 layer model. The architecture of the EpilNet model is shown in Fig. 3.

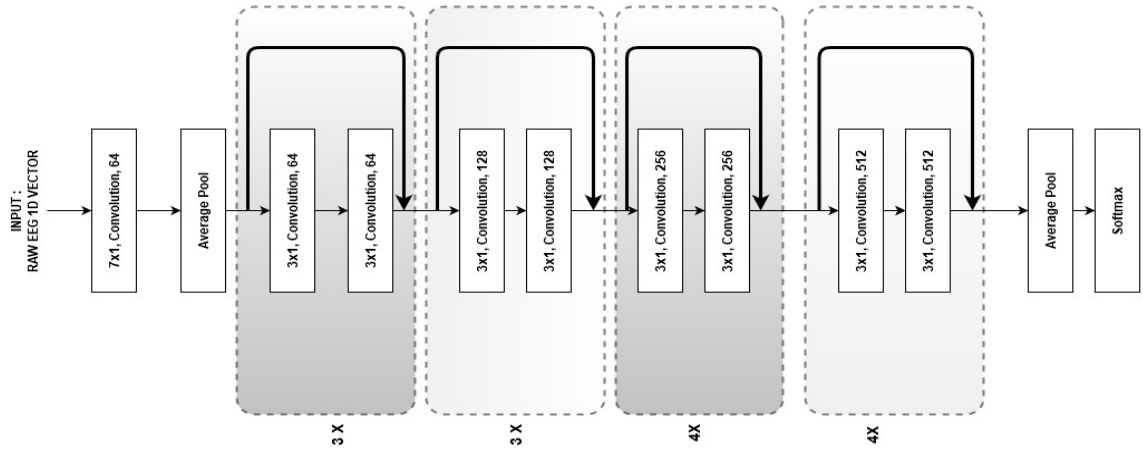

**Fig. 3.** EpilNet model architecture

*2.4. EPILNET ALGORITHM*

The algorithm will help in deciding values of hyper parameters like stride, filter size and dropout ratios in layers of EpilNet Architecture.

---
**Algorithm I:** procedure for basic block used in EpilNet
**Input:** Input vector V, Count of Neuron N.
**Output:** Feature map F is returned.

**procedure** BasicBlock(V, N)
**1:** for i=1 to 2:
**2:**   Convolution operation is applied using filter size 3x1 and stride value 1
**3:**   Batch Normalization is carried out on the obtained vector in the previous step to produce V'
**4:**   Apply non linear activation function ReLU given by f(x) = max(0,x) where x ∈ V' with filter 1x1 and stride 1.
    // To avoid the problem of vanishing gradient use step 5
**5:** Add input vector V to output obtained from steps 1 to 3 to obtain V".
**6:** Apply ReLU f(x) = max(0,x) where x ∈ V" to get F with filter size 1x1 and stride 1
**7: return** F   // return the so obtained feature map F

**end procedure**

---
**Algorithm II**: procedure for EpilNet
**Input:** Signal Vector containing EEG values S, Count of classes C.
**Output:** Class prediction P.

**procedure** EpilNet(S, C)
**1:** Convolution operation is applied on S using filter size 7x1, stride value 2 and padding value of 3
**2:** Batch Normalization is carried out on the obtained vector in the previous step to produce S'
**3:** Apply non linear activation function ReLU given by g(x) = max(0,x) where x ∈ S' with filter 1x1 and stride 1.
*// To reduce computational complexity, overfitting and number of hyper-parameters of model use step 4*
**4:** Apply max pooling operation of filter size 3x1 with stride of 2 and padding of 1 to give output M.
    // Now on the obtained vector M will call BasicBlock procedure with appropriate neurons.
**5:**   for i=1 to 3:
**6:**     M = BasicBlock(M, 64)
**7:**   for i=1 to 3:
**8:**     M = BasicBlock(M, 128)
**9:**   for i=1 to 4:
**10:**    M = BasicBlock(M, 256)

**11:**   for i=1 to 4:
**12:**    M = BasicBlock(M, 512)
**13:**   Apply the average pooling on the final M vector obtained with stride 1.
**14:**   Convert the model network to a dense network with fully connected 512 neurons.

**15:** Reduce the count of neurons in fully connected network to classes count C to give vector P'.
**16:** Apply softmax function $e^{p_i}/\sum e^{p_i}$ where $p_i \in$ P' to obtain probability of each class.
**17:** Predicted class P is the class with maximum probability.
**18:** return P   // predicted class for signal vector S
**end procedure**

The EpilNet is coded in python with the help of pytorch module. About 70% of total data, is used for training. The rest of 12% each is used for validation and testing purpose respectively. In total EpilNet is trained on 20 epochs. One epoch is usually a complete traversal of training data. The best model so obtained, is stored for future testing and prediction in a separate. This file contains learned hyper parameters.

*2.5. WEB API FOR EPILNET*

A Representational State Transfer (REST) based cloud web API endpoint is developed for EpilNet. The endpoint is designed using flask module in python. The API helps in bringing EpilNet into practical use. The API endpoint takes in JSON (JavaScript Object Notation) feed as parameters in HTTP POST request. The data parameter contains EEG signal 1D vector collected from IoT device. The API is using 'Gunicorn' web server that serves each request concurrently. The web API endpoint is deployed on herokuapp for public use by developers around the globe. Herokuapp provides balance of performance, flexibility and configuration simplicity.

*2.6. MOBILE BASED PREDICTION SYSTEM USING EPILNET*

A mobile based android application is developed. This will enable prediction system to be anywhere, anytime. The application will take in EEG signal from IoT device. It will help in predicting on-site time of seizure. Whenever a seizure on-site time is predicted then application will generate an alarm to alert patient. Along with the alarm it will also fetch location details of patient. The location details will be sent to emergency contacts using SMS. This will assist caretakers to help patient before any seizure especially in case of small children and pregnant ladies. When an actual seizure takes place at that time, application will send location to emergency contacts. Along with that the application will alert hospital if seizure time exceeds five minutes. The system will also send details about seizure to doctor using an email. The information about seizure will be stored in cloud database to assist easy and efficient diagnosis by doctors using proposed web portal in next section.

*2.6.1. Architecture and flowcharts of application*

The architecture of application is shown in Fig.4. Basically the dataset is used for training EpilNet. The trained model is then saved in external file. The cloud API component helps in brining EpilNet in to real world use. The android service component takes in input signal from IoT device and sends it to cloud API for testing using internet service. Whenever a pre-ictal is predicted as result then alarm service component is used to alert patient. Also along with that SMS service component is used to send SMS to emergency contacts about current location details. These location details are fetched using network based component or GPS based whichever is available. Similarly when an ictal is predicted then an email about details will be mailed to doctor using email service component. Along with that in ictal prediction, SMS service component is also used.

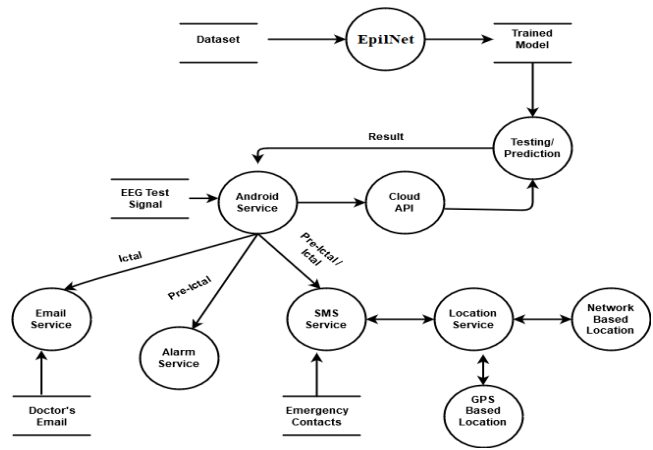

**Fig. 4.** Architecture of application

The flowcharts for working of components of mobile application, is shown in the Fig. 5. The components shown in flowchart include starting a service, stopping a service and managing contacts. Managing contacts involve all aspects starting from adding new emergency contacts, view saved contacts and deleting any existing contact. The data or conditions required are shown in flowcharts for each component.

*2.6.2. Distinguish features of mobile application*

The additional distinguish features that present mobile application has over the existing works, are listed below:

- Less resource utilization.
- Better battery optimization - As application sleeps down during inactive period and only wakes up when it needs to send input signal and take action accordingly.
- Requires less computational power – Application is active only for 1 minute out of 5 minutes slot, thus leaves computational power 4 minutes free for other applications.
- Easily switching between network-based location and GPS based location, for better availability of the location service.
- Protecting the service from being killed by CPU with use of foreground service instead of daemon services.
- Prediction service requires a minimal user interaction.
- Integrated alarming and prediction system for both patient and doctor.
- Better usability and operability.

*2.7. WEB PORTAL FOR DIAGNOSIS*

The web portal is developed that will assist the doctors at hospitals, in making treatment and analysis process efficient, faster and reliable. The main feature of portal is to reduce the time taken to predict epilepsy using EEG graphs. The manual process is not only cumbersome but also requires a deep knowledge and experience. EpilNet will help doctors to take signals from EEG device and easily predict whether patient is healthy or unhealthy. EpilNet used for portal is trained for all five classes. Since EEG graphs used in manual prediction are generally of low quality. The portal will also assist doctors in better studying of EEG graphs. It will plot digital EEG graphs for input signal taken from EEG device. These graphs could be scaled and zoomed upto milliseconds for studying of minute details. This will help doctors in better analyzing the current status of patient and choosing among available treatments wisely.

Only for registered patients portal will provide his/her history like frequency of seizure, time etc. Doctors just has to input patient id and all the details of past seizures will be displayed. These details are captured and stored using mobile application described in earlier sections. This will help doctor decide the appropriate dosage pattern for patient. Along with that it will help doctors examine whether the past medication is showing any improvements in patient.

## 3. Experimental Results

In the present study, EpilNet is using 20 epochs for training purpose. The trained model is further used for development of web API, mobile application for patients and web portal for doctors at hospitals. The complete detailed analysis of observations and results acquired during the development are given below.

*3.1. EPILNET MODEL EXECUTION*

**Group 1:** To classify between the healthy brain activities (A, B) from pre-ictal (D) and ictal (E).

The EpilNet model accuracy during training and testing session, is shown in the Fig. 6. The best model obtained during training phase, is saved in external file. In group 1, the best fit model is obtained at 12th epoch. The highest training accuracy so obtained is 95.18%. The testing accuracy for the best fit model obtained, is 94.55%.

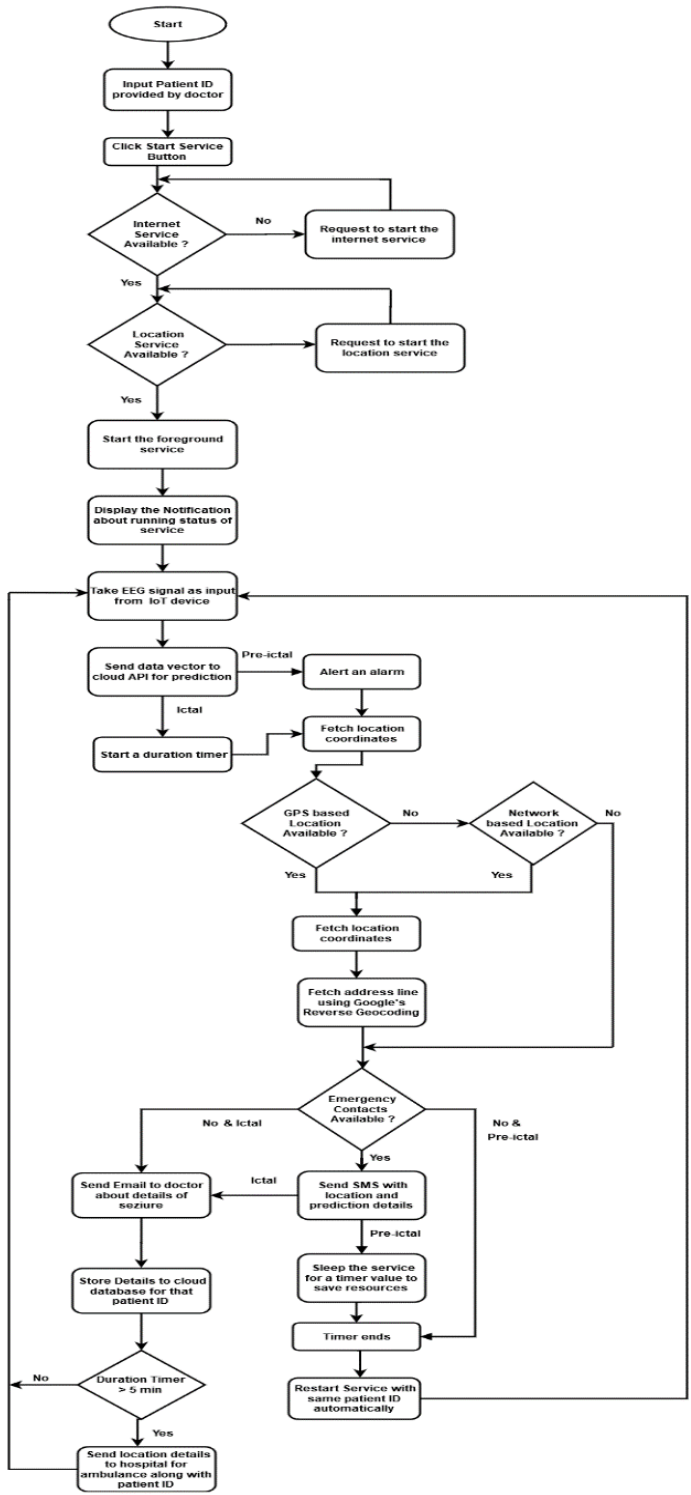
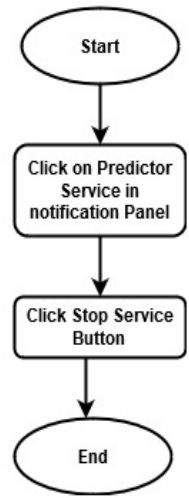
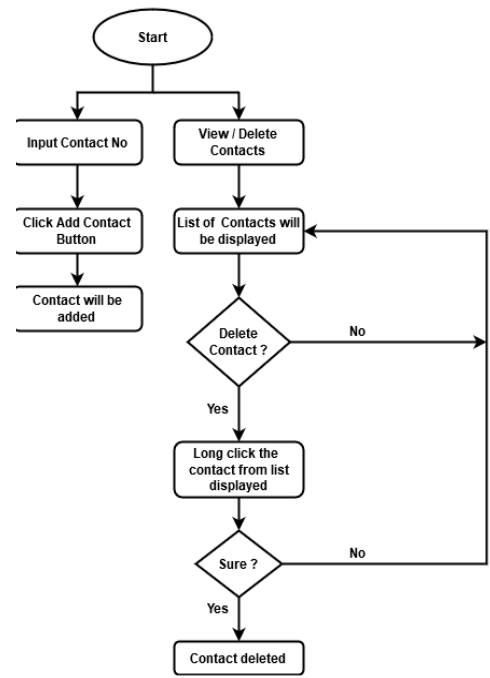

**Fig.5.** Flowchart for working of application (a) Starting Service (b)Stopping Service (c) Managing Contacts

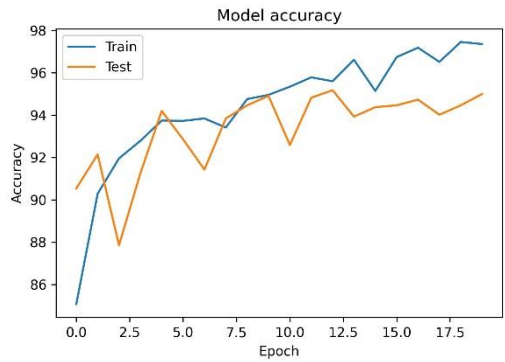
**Fig. 6.** Training v/s. Testing accuracy for group 1

Confusion matrix generated while performing Testing session is shown in Fig.7.

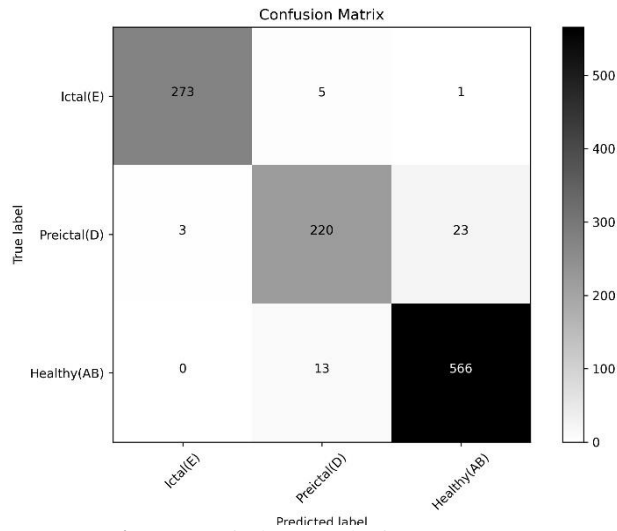
**Fig. 7.** Confusion matrix for group 1

**Group 2:** To identify between all the five classes

The EpilNet model accuracy during training and testing session is shown in the Fig. 6. The best model obtained during training phase is saved in external file. In group 2 the best fit model is obtained at 14th epoch. The highest training accuracy so obtained is 82.58%. The testing accuracy for the best fit model obtained is 79.13%

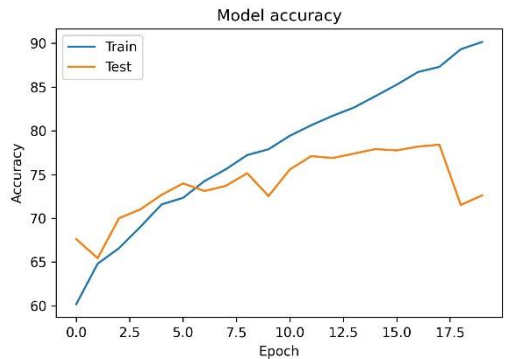
**Fig. 8.** Training vs. Testing accuracy for Group 2

Confusion matrix generated while performing Testing session is shown in Fig.9.

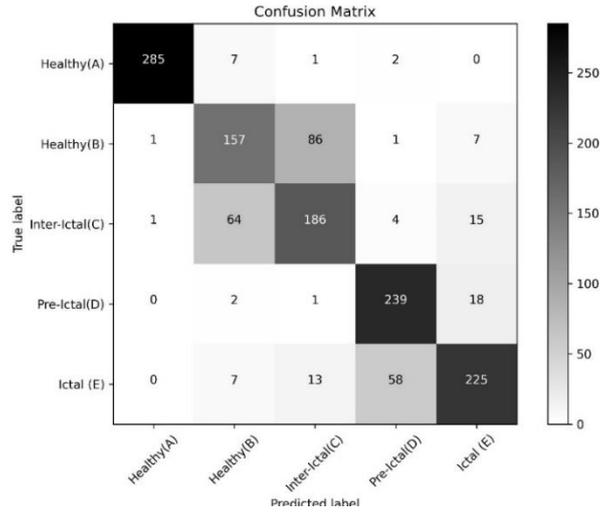

**Fig. 9.** Confusion matrix for group 2

*3.2. PERFORMANCE EVALUATION OF WEB API*

EpilNet deployed on herokuapp using 'Gunicorn' server. The server uses a worker that serves the incoming request. To better analyze the performance of web API load testing is carried out. Apache jmeter is used for load testing. A parallel load 100 registered patients is used for testing. Testing is performed for all the five classes. The load of 100 patient request is maintained for 60 seconds with a startup time of 10 seconds. The graph for response time (in milliseconds) against timestamp is shown in Fig. 10.

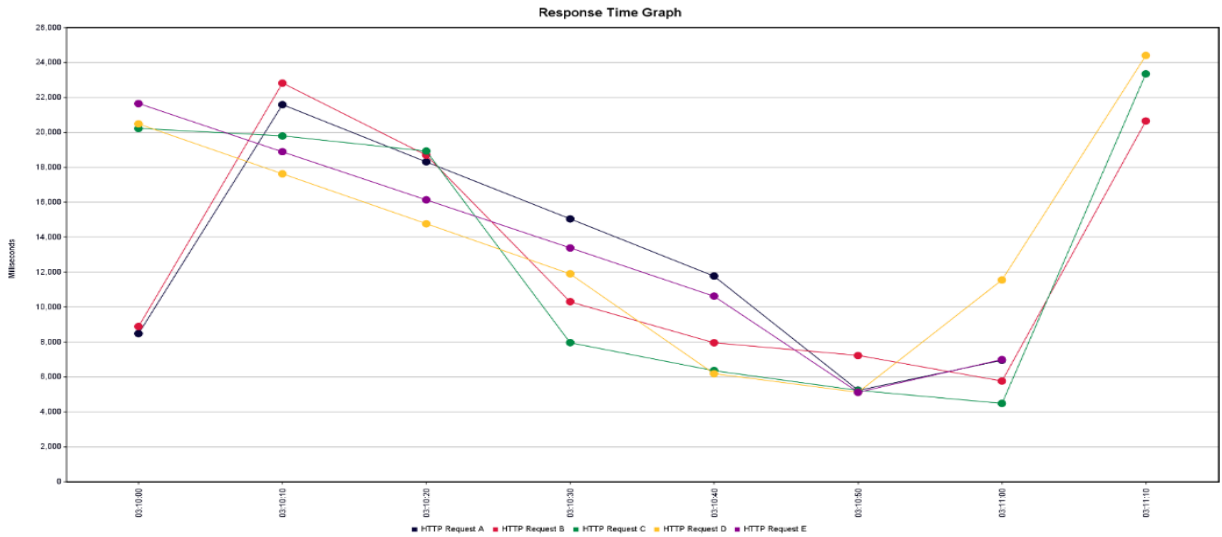

**Fig. 10.** Response time graph for web API

The summary report of response time (in milliseconds) of web API, is presented in Table 2. For each request class a sample of 100 parallel load is used for generating the summary report.

**Table 2.** Summary report of response time (in milliseconds)

| Label | Samples | Average | Min | Max | Std. dev. | Throughput |
|---|---|---|---|---|---|---|
| Request A | 100 | 13022 | 1578 | 25376 | 7690.73 | 1.5 /sec |
| Request B | 100 | 12997 | 4119 | 25049 | 6675.90 | 1.2 /sec |
| Request C | 100 | 11907 | 4169 | 24760 | 8075.66 | 59.2 /min |
| Request D | 100 | 5944 | 4332 | 24644 | 3743.19 | 56.9 /min |
| Request E | 100 | 5303 | 3869 | 7522 | 842.61 | 2.9 /sec |

*3.3. SNAPSHOTS OF MOBILE AND WEB APPLICATION*

The various snapshots for the mobile application are presented in Fig. 11. For More clarity of the working of proposed application.

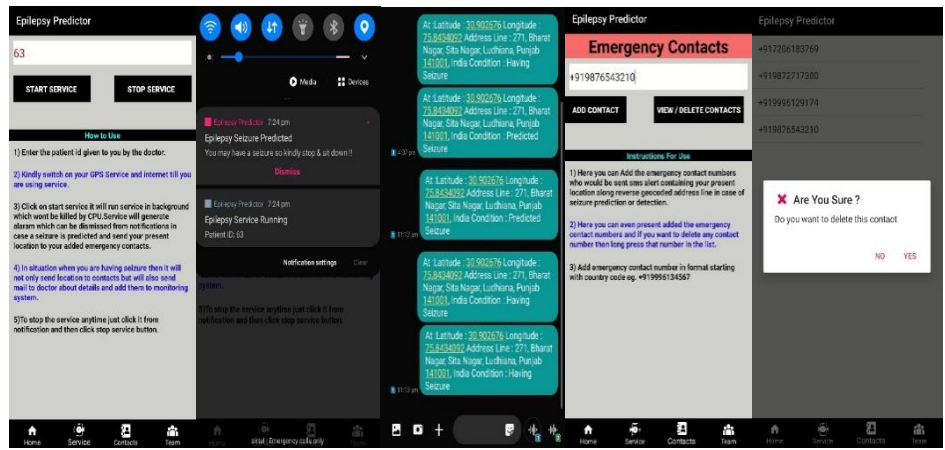

**Fig. 11.** Snapshots of application

The web portal which is used at hospitals, is shown in the Fig. 12.

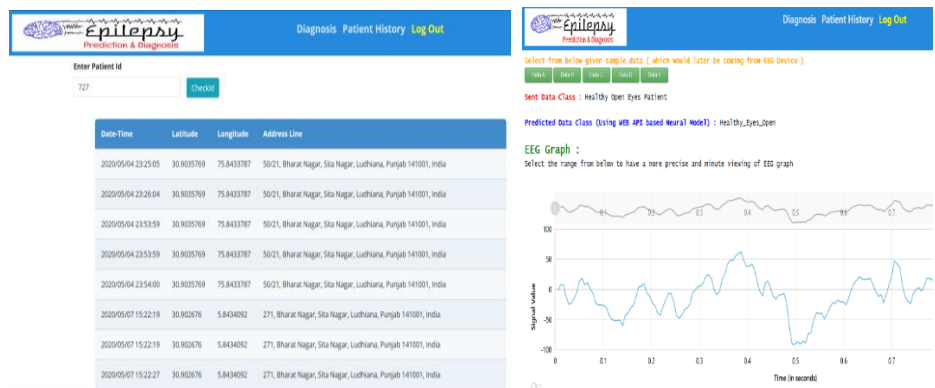

**Fig. 12.** Snapshots of web portal

## 4. Discussion

Liu et al. implemented a 2D convolutional neural network (CNN) which was fed with multi-biosignals [13]. These multi bio-signals combined EEG, electromyography (ECG) and respiratory signals. The CNN achieved an accuracy of 65% for five classes. Mao et al. combined the continuous wavelet transform (CWT) and CNN to classify epileptic seizure [14]. The experiment used the wavelet transformation for conversion of EEG data into time frequency domain images. These images were fed into a 2D CNN which consisted of three set of convolution, activation and pooling layer. The 2D CNN achieved an accuracy of 72.49%. Abbasi et al. proposed Long Short-Term Memory (LSTM) based classifier [1]. In the preprocessing, Hurst Exponent and Autoregressive Moving Average (ARMA) features are extracted from each signal. The extracted feature are then used by double layered LSTM for classification. For classification into signals of three kind it achieved an accuracy of 95%. For binary classification into inter-ictal and ictal, the accuracy increased to 98%. EpilNet achieved a superior performance compared to other studies in literature as shown in Table 3. EpilNet model has following advantages:
- Preserves the true textual nature of raw EEG data.
- Overcomes the problem of vanishing gradient.
- Achieves a testing accuracy of 79.13% for five classes, highest when compared to present works.
- EpilNet also performs better for classification into three classes achieving accuracy of 94.55%.
- Can be easily brought into practical use.

Cruz et al. developed a standalone mobile application [10]. The application predicted the ictal state (When patient is having seizure) using wearable glove. The glove used accelerometery and electromyography (ECG) as measurement signals. The application used SVM algorithm and raised a SMS alert only. Rangathan et al. proposed a seizure diary for patients to record seizure activity manually [6]. Marzuki et al. developed an application 'MyEpilPal' [15]. It is a self-management and monitoring tool that can support the epilepsy patient and the caregiver. The application allows patient to manually add details about seizure. Also application helps in manually storing dosage of medicine. This enables monitor side-effects and effectiveness of antiepileptic medicine. The proposed integrated system consisting of Web API, Android application and web portal has following advantages:

- Automated Internet of Things (IoT) enabled platform. It predicts on-site seizure time, raises an alarm alert, SMS alert to caretakers with location and also stores seizure details on cloud database for later effective treatment by doctors at hospital using web portal.
- Android application has better resource utilization, requires less computational power and battery.
- Easily switches between network based and GPS based location, for better availability of location service.
- Web portal also helps in faster diagnose of new patients into five classes of medical significance at hospitals.
- Web API has average response time of less than 25 seconds.
- Easy operability and better usability.

## 5. Conclusion and Future Work

In the present study, EpilNet model is proposed. It overcomes the problem of vanishing gradient. Basic blocks of the model is fed with additional skip connection to reduce the vanishing gradient along with retaining the true textual nature of EEG signals. With this, EpilNet gives the testing accuracy of 79.13% for five classes leading to significant increase of about nearly 6-7% when compared to related works. EpilNet gives 94.55% for classification between healthy, pre-ictal and ictal. Further, the developed web API helps in bringing EpilNet into practical use. API has an average response time of less than 25 seconds for all the five classes. The integrated system proposed predicts the on-site time for seizure quickly. This helps patient be safe. Also developed system helps doctors in better and faster analysis of patients at hospitals, thus increasing efficiency of treatment process. The developed application not only has lesser resource utilization but is also more reliable, usable and easy to operate. The limitation of the system is that response time of web API can be further reduced. It can be achieved by further optimizing the concurrency and parallelism level.